\documentclass[runningheads]{llncs}

\usepackage{makeidx}
\usepackage{graphicx}
\usepackage{amsmath,amssymb} 
\usepackage{color}
\usepackage[ruled,vlined]{algorithm2e}
\usepackage{subfiles}
\usepackage{authblk}

\begin{document}
	\pagestyle{headings}
	\mainmatter



	\title{Contact Area Detector using Cross View Projection Consistency for COVID-19 Projects}

	
	
    
    
    
    \author{Pan Zhang\inst{1} \and
    Wilfredo Torres Calderon\inst{2} \and
    Bokyung Lee \inst{1}
    Alex Tessier\inst{1} \and
    Jacky Bibliowicz\inst{1} \and
    Liviu Calin\inst{1} \and
    Michael Lee\inst{1} 
    }
    %
    %
     
    
    \institute{Autodesk Research\\
    \email{\{pan.zhang, bokyung.lee,Alex.Tessier, Jacky.Bibliowicz, liviu.calin, michael.lee\}@autodesk.com}\\
     \and
    University of Illinois, Urbana-Champaign\\
    \email{trrscld2@illinois.edu}}

	\maketitle

\begin{abstract}

The ability to determine what parts of objects and surfaces people touch as they go about their daily lives would be useful in understanding how the COVID-19 virus spreads.
To determine whether a person has touched an object or surface using visual data, images, or videos, is a hard problem. Computer vision 3D reconstruction approaches project objects and the human body from the 2D image domain to 3D and perform 3D space intersection directly.  However, this solution would not meet the accuracy requirement in applications due to projection error. Another standard approach is to train a neural network to infer touch actions from the collected visual data. This strategy would require significant amounts of training data to generalize over scale and viewpoint variations. 
A different approach to this problem is to identify whether a person has touched a defined object. In this work, we show that the solution to this problem can be straightforward. Specifically, we show that the contact between an object and a static surface can be identified by projecting the object onto the static surface through two different viewpoints and analyzing their 2D intersection. The object contacts the surface when the projected points are close to each other; we call this {\em cross view projection consistency}.
Instead of doing 3D scene reconstruction or transfer learning from deep networks, a mapping from the surface in the two camera views to the surface space is the only requirement. For planar space, this mapping is the Homography transformation. This simple method can be easily adapted to real-life applications.
In this paper, we apply our method to do office occupancy detection for studying the COVID-19 transmission pattern from an office desk in a meeting room using the contact information. 

\end{abstract}

\section{Introduction}

COVID-19 virus transmission occurs, in part, from touching contaminated objects or surfaces. The ability to determine when and where a person touches an object in space is essential in understanding the spreading path of the virus and helping reduce the spread (for example, helping facilities staff to prioritize their cleaning). A fast and accurate method for contact tracking is both ideal and urgent to curb the spread of COVID-19. Computer vision techniques are ideal tools for automating this process and generating insightful statistics, as they are efficient, accurate, adaptable, and video sensors can be placed safely and unobtrusively in the space.

The task of answering whether a person has touched something is challenging but has been solved. The solution has been actively applied to automated grocery store products like Amazon Go. However, the bar of implementing such a system is still unreachable for everyday use cases without the support of a team of engineers and researchers when lots of sensing technologies need to work together with customized algorithms. This paper focuses on solving this problem using only two commodity cameras with simple algorithms and little calibration involved. Our method gives a solution that can be made ubiquitous and implemented inexpensively. 

Two straightforward approaches might solve this problem. The first one is to reconstruct 3D human joints and the object from 2D camera captures; then computing the touched areas from the 3D intersections between human end joints and the object. With this approach, given a correctly calibrated camera system, errors could still be introduced during 3D reconstruction steps. Furthermore, the reconstruction requires multi-view object identification, which is still an open research area, and this adds an extra challenge to the task. The errors introduced by these issues give inaccurate 3D reconstruction results, which produce inaccurate intersection results. 

The second approach is to train a neural network to detect human touch actions. The network could be trained to do a regression task in both spatial and temporal domains. If done correctly, a satisfactory result could be achieved using a single-camera-setup. However, the challenges, such as creating a large scale human-to-object contact dataset and methods that can generalize its application, make this deep learning approach very hard to implement in the short term to address urgent application needs. 

\begin{figure}[t]
    \centering
    \includegraphics[width=0.99\columnwidth]{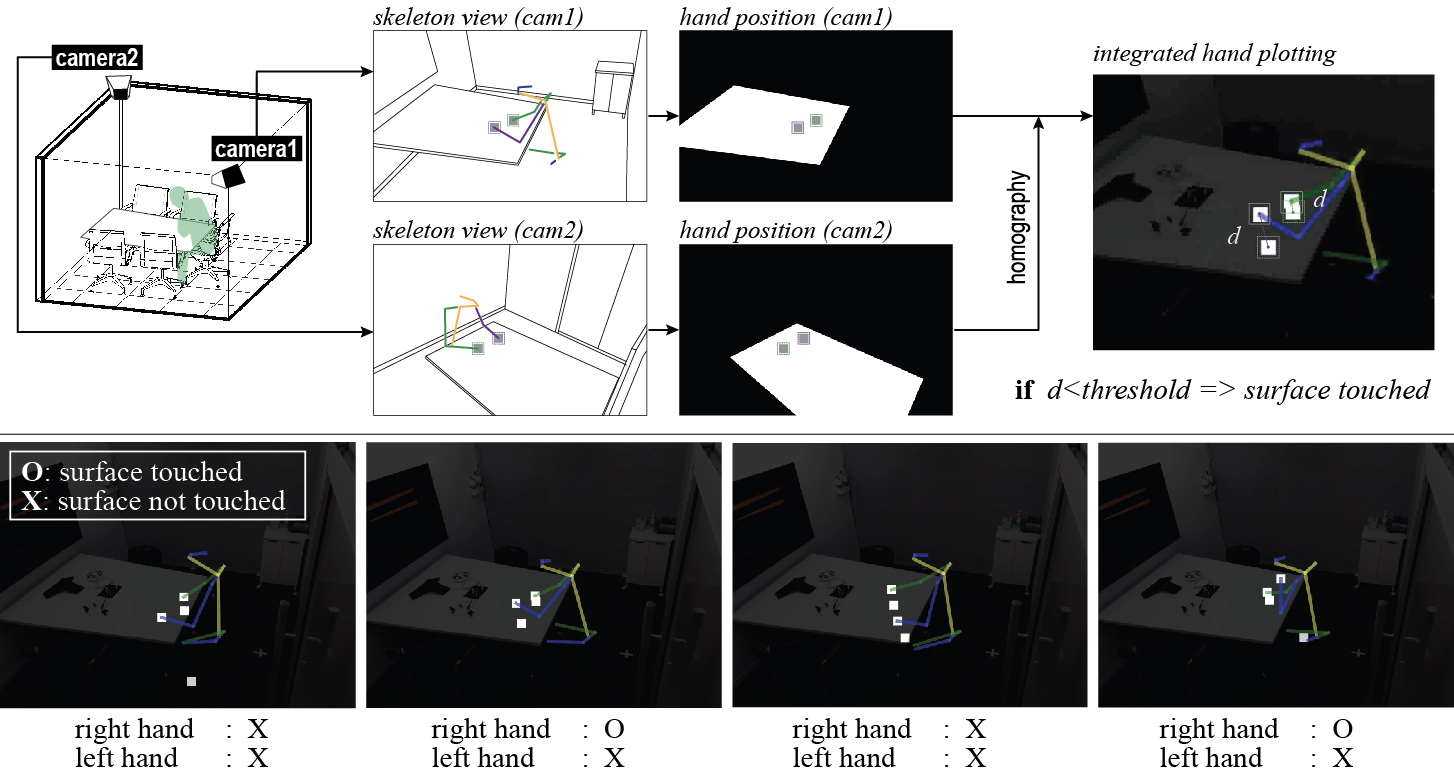}
        \caption{Overview of the method.}
        \label{fig:teaser}
    \end{figure}
However, this question is the same as to know whether a person has touched a defined object. Furthermore, we show a straightforward method that can solve this contact detection.
Particularly, for a surface that can be viewed from two different viewpoints, we analyzed the projection of the two viewpoint images to the surface plane. Assuming there is only one object in the scene, if the object's projection from both camera views intersects on the surface plane, the object is considered in contact with the surface. We call this {\em cross view projection consistency}, as shown in Figure \ref{fig:teaser}.

We applied this method to determine the contact area for a desk in an office meeting room. Since the desk surface is planar, we can use Homography mapping to map between surfaces. Furthermore, we extend our method to find the mapping for non-planar surfaces. Because of the uniqueness of the mapping, one can create a mapping from a recorded video with an object moving on the surface until the track of it covers the whole surface. Then, the point of the object center being detected in the images from the two cameras captured at the same time becomes the map of the surface point between the two camera views; we call this surface mapping calibration.  

We use the contact information gathered from the method in a case study to understand the COVID-19 spread pattern. In particular, we generate a contact heat-map on the desk surfaces in the office meeting rooms and occupancy detector as in Fig \ref{fig:occupancy} to show the usage frequency of the desks as shown in Fig \ref{fig:heatmap}. 

In summary, our contributions are: i) a simple accessible method that finds the contact area using commodity hardware and easy setups; ii) an easy mapping calibration method for non-planar surface; iii) an application of our method that might be used for the study of COVID-19 transmission patterns in the office.   


\section{Related Work}

	\subsubsection{Human object contact area detection: }
	
	Human-computer interaction (HCI) has played a significant role in developing interactive means to improve and smooth communications between computers and humans. The evolution from keyboards to interactive projection surfaces to obtain human input is an example that highlights its progress. However, it relies on more sophisticated software and hardware (e.g. stereo \cite{agarwal2007high,matsubara2017touch} and rgb-d \cite{wan2015explore,linqin2017dynamic} cameras) to acquire the data. Agarwal \cite{agarwal2007high} proposed machine learning and geometric models to capture finger-surface contact using an overhead stereo camera. Matsubara et al. \cite{matsubara2017touch} reasoned about the shadows produced by the fingers when hovering or touching a surface. They presented a machine learning algorithm using the input from an infrared (IR) camera and two IR lights to identify finger-surface contact. The proposed strategy in this project relies on a simple but effective way to identify contact by leveraging the availability of standard monocular cameras typically used in office settings.
	
	\subsubsection{Action recognition based approach: }
	
	Vision-based action recognition is focused on computing robust action features for the correct identification of human actions. Different from action classification \cite{carreira2017quo,yan2018spatial}, action detection or action segmentation \cite{lea2017temporal,farha2019ms,chen2020action} carries an additional level of difficulty by trying to identify the start and ending time of an action in the temporal dimension. Carreira \cite{carreira2017quo} presented the advantages of using a pre-trained model for action classification. Yan \cite{yan2018spatial} explored the temporal component using body pose key-points as input through a Spatial-Temporal Graph Convolutional Network (ST-GCN) for the same task. The approach for the temporal location of actions introduced by Farha \cite{farha2019ms} implemented a Multi-Stage Temporal Convolution Networks (MS-TCN) that refines its prediction on every stage to classify and find the starting and ending time of each action. The work of Chen \cite{chen2020action} extended this idea by incorporating self-supervised domain adaptation (DA) to the MS-TCN network. The objective of these approaches differs from ours in that action recognition does not require identifying the location of the surfaces touched by a human, but understanding the sequence of movements of a person's appearance or limbs with the labeled activity.

	\subsubsection{3D pose estimation based approach: }
	
	3D Human pose estimation of single-person and multi-person in individual frames and videos continue to attract significant attention \cite{wei2016convolutional,bulat2016human,pavllo20193d,kolotouros2019learning}. Large-scale benchmark datasets and deep learning models have made successful progress \cite{ionescu2013human3}. However, the large variety of human poses, occlusion, and fast motions still represent significant challenges for this research area. State-of-the-art (SOTA) methods use 2D key-point (e.g., wrists and ankles) locations as input paying less attention to the position of the palm and fingers, central parts for human-to-surface contact analysis. Even with successful identification of the 3D human pose, complete contact detection requires 3D scene understanding. Whether using a precomputed 3d reconstruction or a 3D CAD model, both would need to be continuously updated when movable objects (e.g., furniture) change locations. These limitations and the intrinsic ambiguities of 3D human reconstructions would affect the accurate identification of surface-human contact.

\section{Methodology}

  
    In this section, we first introduce \textit {cross view projection consistency}, which is the theory behind our method. Then we show an informal proof of it to illustrate the correctness and show that we can relax the assumption, in theory, to use in our case. In section \ref{subsection:surface_contact}, we describe the algorithm we use for desk surface contact recovery. In section \ref{subsection:surface_mapping}, we introduce the surface mapping calibration method for finding mappings of the non-planar surface. 
    
    \subsection{Cross view projection consistency and its limitation: } \label{subsection:surface_contact}

    \subsubsection{Cross view projection consistency:} Given a surface that can be observed from two cameras at different locations, assume a small ball exists in the space. When the ball overlaps with the surface in both camera views, we can project the ball's location onto the actual (3D) surface to compute the intersection points.  The intersections will be closest to each other if and only if the ball is in contact with the surface, and we call this projection consistency.

    \subsubsection{Informal proof for illustration purpose:} . As a general surface can be decomposed to small planar patches, the proof will imply the correctness for all surfaces if we illustrate this to be true for a planar patch. For a planar patch or a planar surface, simple observations can prove cross-view projection consistency. As shown in Fig.2. Given a planar surface, a spherical object, such as a basketball, and two spotlight sources that illuminate the area from different directions that are not perpendicular to the surface normal, the ball has the smallest shadow area only when it touches the ground. The shadow will run away from each other them the ball goes up in the air. If we replace the spotlights with cameras, the shadows on the surface are the maps of the intersections. 
    
    \begin{figure}[t]
    \centering
    \includegraphics[width=0.8\columnwidth]{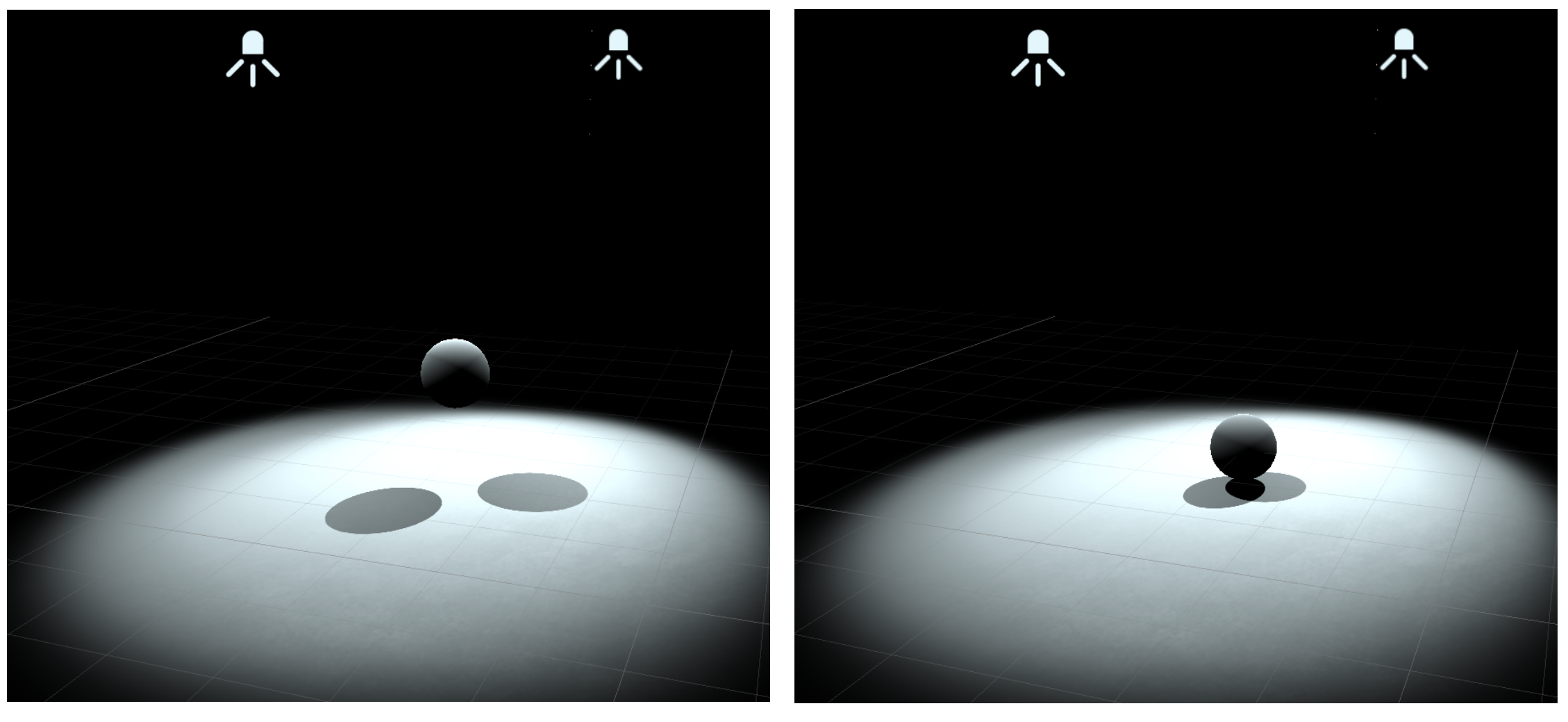}
        \caption{Shadow changes with respect to the object's distance from the surface.}
        \label{fig:ball-shadow}
    \end{figure}
    
    \begin{figure}[t]
    \centering
    \includegraphics[width=0.99\columnwidth]{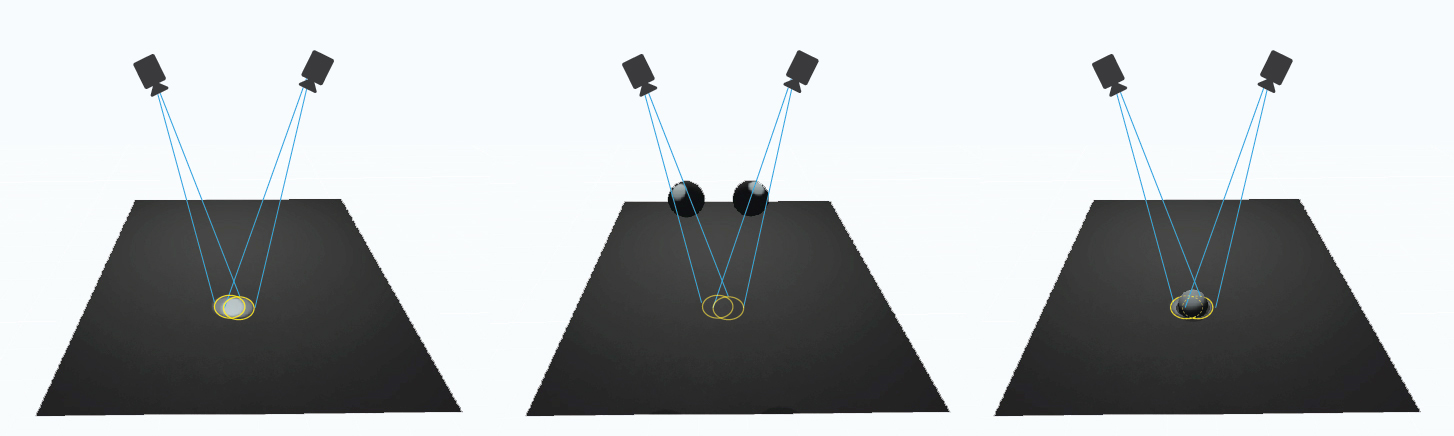}
        \caption{
        Multi-view perspective of a single object}
        \label{fig:single-object}
    \end{figure}

    \subsubsection{Single object assumption:}
    
    \textit {Cross view projection consistency} holds when there is only a single object in the scene. Having more than one object in the scene causes ambiguities, as shown in Fig.3, unless we have a way to identify the same objects from different camera views, which is still ongoing research. Without doing camera calibration and 3D reconstruction or machine learning-based texture mapping, it is challenging to solve. However, this assumption might be relaxed given the condition that the camera is far away relative to the object, which is the same as to say the object is small enough in the scene.  We found that under this condition, the chances of ambiguities being created by more than one object are minor. Here, we use a toy example to demonstrate the probability of ambiguity is very low, again, as the formal proof requires lots of references from radiometry theories that are not in the scope of this paper, and we use loose terms only for demonstration purposes. Assume $2$ freely moving objects in the scene that can be viewed by $2$ cameras; then, for a given small patch on the surface, there is only one cluster of rays from each of the camera needed for this patch to be visible in the camera view. The ambiguity happens only when the two objects are blocking the two clusters of rays simultaneously to make the patch not visible from any of the camera view. The probability of occurrence is the volume of two clusters of the rays over the volume of the space, which is close to zero when the object is small in the view.  We claim that to detect hand-object contact, this condition generally holds, as people's hands are small in the image and can be modeled by points or small patches.
    
    \subsection{Homography mapping for planar surface: } \label{subsection:surface_mapping}
    Using \textit {cross view projection consistency}, this section shows the algorithms of finding contact areas on the desk. Since the desk surface is planar, we use a Homography to map points surface between different views. Assume a two-camera setup, a plane $\alpha$, an object $m$ at point $p$ in 3D space, at a time $t$ when $m$ can be viewed by both cameras. The context above to can be translated to this formulation. 
    \[  p_x = F_d (p_1 , \mathcal{H}(p_2)) \]
    where $p_1$ is the center of the object in camera $1$ view and $p_2$ is the center of the object in camera $2$ view,  $\mathcal{H}$ is the Homography mapping that maps the plane $\alpha$ from camera $2$ view to camera $1$ view, and $F_d$ is the consistency checking function that returns $p_1$ if the distance between $p_1$ and $p_2$ is smaller than $d$ and returns empty otherwise. And $p_x$ is the location on the plane $\alpha$ being contacted by object $m$ at time $t$ from camera $1$ view.
 
    Given the formulation above, we use algorithm 1 for detecting the contact area on the surface for the implementation.
    
    \begin{algorithm}[H]
    \SetAlgoLined
    \KwResult{Set $Q$ of mappings that maps timestamp to contact points }
     $Q = \emptyset$ \;
     \For { t in all time stamps }{
        create an empty point set $P = \emptyset$ \;
        \For { object center $p_1$ in the centers detected from camera $1$ }{
        
                   \For { object center $p_2$ in the centers detected from camera $2$ }{
                               
                                   $p_x = F_d (p_1 , \mathcal{H}(p_2))$ \;
                                   $P=P \cup p_x$ \;
                                   $Q=Q \cup (t\mapsto P$) \;

                      }
           }
     }
     \caption{Contact detection}
    \end{algorithm}
    
 
    \subsubsection{Noise from the detection: }
    The implementation of the contact area detection algorithm requires objects to be detected from both camera views. The detection algorithm can be chosen from object joint detection \cite{fang2017rmpe,cao2018openpose}, instance segmentation \cite{he2017mask}, optical flow detection \cite{lucas1981iterative,suhr2009kanade}, or by just computing the intensity variance pattern on the surface. We model the detected object center $p_x$ as $p_x \sim \mathcal{N}( p_\mu,\,\sigma^{2})$ . 
    where $p_{\mu}$ is the real object center, and $\sigma^{2}$ is the variance. We include a plot to indicate the different std value might change the result by show plots of smallest distance between $p_1$ and $\mathcal{H}(p_2)$ overtime as shown in Fig.4.
    
    \begin{figure}[t]
    \centering
    \includegraphics[width=0.7\columnwidth]{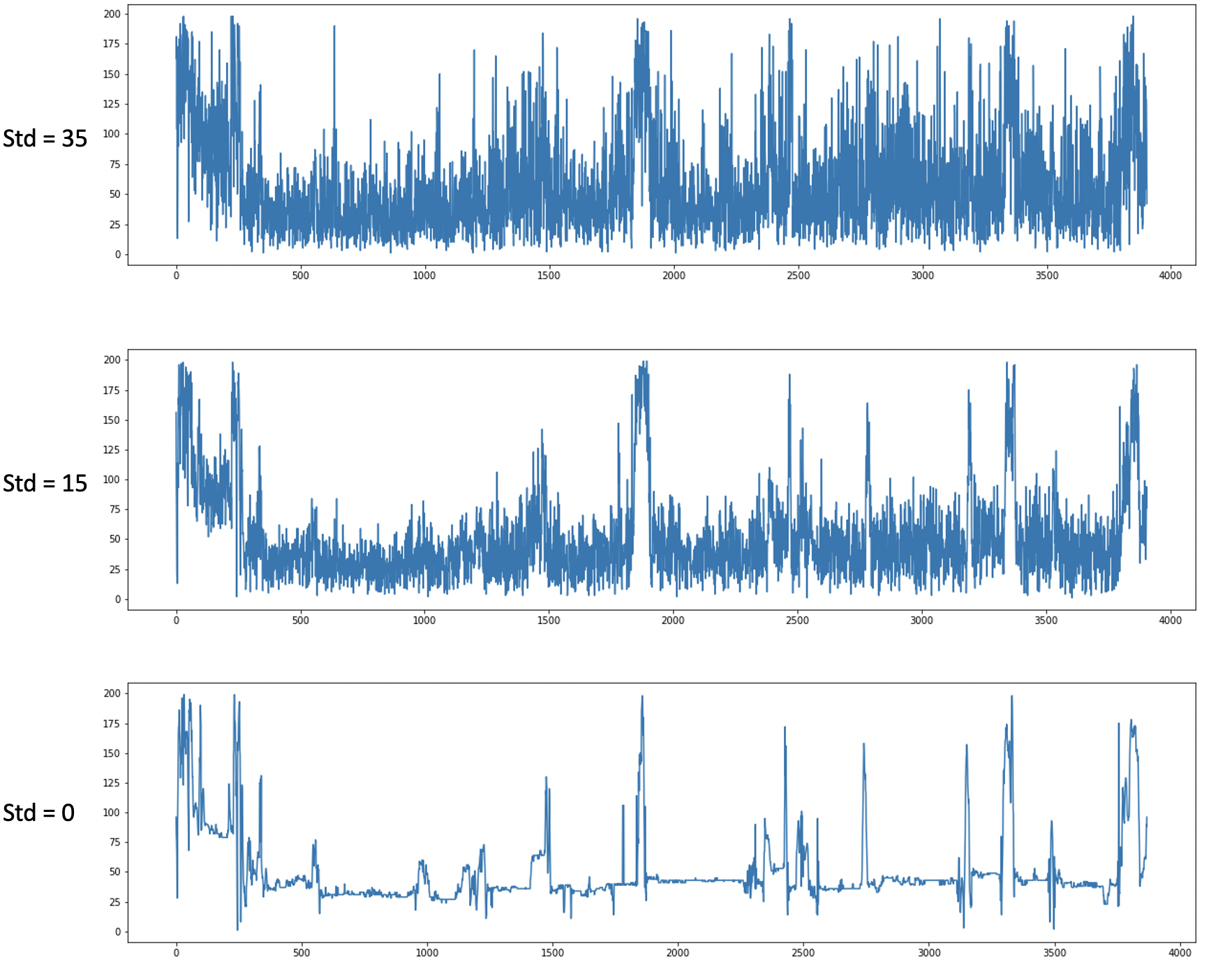}
        \caption{
        Figures that shows the distance between $p_1$ and $\mathcal{H}(p_2))$ of the wrist detected by adding different noise models to the detection, the unit on $y$ axis is per pixel, the unit on $x$ axis is per 0.04 second 
        }
        \label{fig:example}
    \end{figure}

    \subsubsection{Set distance threshold: }
    Every object has a volume, so even for the case of a perfect contact on the surface with no error introduced when computing centers of the object from the two cameras and with no error introduced in the Homography mapping,  the two centers $p_1$ and $\mathcal{H}(p_2))$ will not be the same. We need to set $d$ parameter for the consistency checking function $F_d$.
    
    Parameter $d$ needs to be tweaked according to the scene setup but once setup, it does not need to be changed. One simple way is to do this by observation. One can select a clip of video when contact happens and compute the $d$, which is the Manhattan distance between points in the view of camera $1$ and the points in the Homographically transformed view of camera $2$.
    
    Another more accurate way of doing this is to perform calibration when the person has access to the scene; one can go into the scene, keep his hand visible, contact the desk, and swipe the entire desk surface. From this video, a map from patches to the $d$ threshold when contacting happens is calculated. This follows a very similar thought as \textit{surface mapping calibration} introduced below. Explicitly, in our study for the office desk contact area detection shown in the applications section, we compute a $d$ map for each point on the desk surface in the camera $1$ view for the entire desk area. In this setup, we rasterize the desk surface into small patches. Then, we select videos with only one person in the scene (the videos are selected using human tracking result). From the video, for each wrist joint that falls into a patch in the 2D image, we collect the distance between the joint point in camera $1$ view and the joint point in camera $2$ being Homographically transformed to camera $1$ view. Doing this process over time, we will have a bag of distance numbers collected for each patch. Furthermore, we select the $d$ value from these numbers. From the observation, we find that people spend most of the time resting their arms on the desk, and the most values collected should be the moment contact. We can use histogram intervals, and the one interval that receives the most numbers can be used to set the $d$ number for that patch. For the patches that have no number or do not have enough numbers collected, the $d$ value can be interpolated from its neighbor patches. 
    
    \subsubsection{Surface mapping calibration: }
     The method also supports getting the contact area on a non-planar surface. For the non-planar surface, the challenge is to find the mapping of the points on surfaces between $2$ camera views. For the surfaces with rich texture patterns, algorithms like SIFT \cite{lowe2004distinctive} mapping algorithms can be used. However, for the surfaces that have no or low quality texture patterns, we can create the mapping by detecting objects when they are contacting the surface. The intuition behind this is from a structured light system for doing 3D reconstructions. The light source projects patterns on the surface and the camera maps the pattern and knows the map of it on the image, now, when the object contacts the surface, it can just be thought of as a pattern being projected on the surface.

\section{Application to COVID-19 Projects}

    In this section, we show some ways to use our result which might help to understand the COVID-19 spreading pattern. In section 4.1, we give an introduction to our dataset, in section 4.2, we show contact heat-maps of a office desk with data collected over a month. In section 4.3, we introduce desk occupancy detection using from which we believe that estimation of the virus transmission might be computed.
    
    \subsection{Dataset:}
    We use the office dataset from \cite{Lee:cscw:2019}, The people from the video are not performers, they are office users being captured from two cameras as shown in Fig.1. The cameras capture activities in the office during working hours for about 45 workdays. Humans in the video are skeletonized for privacy protection.

    \begin{figure}
    \centering
    \includegraphics[width=0.99\columnwidth]{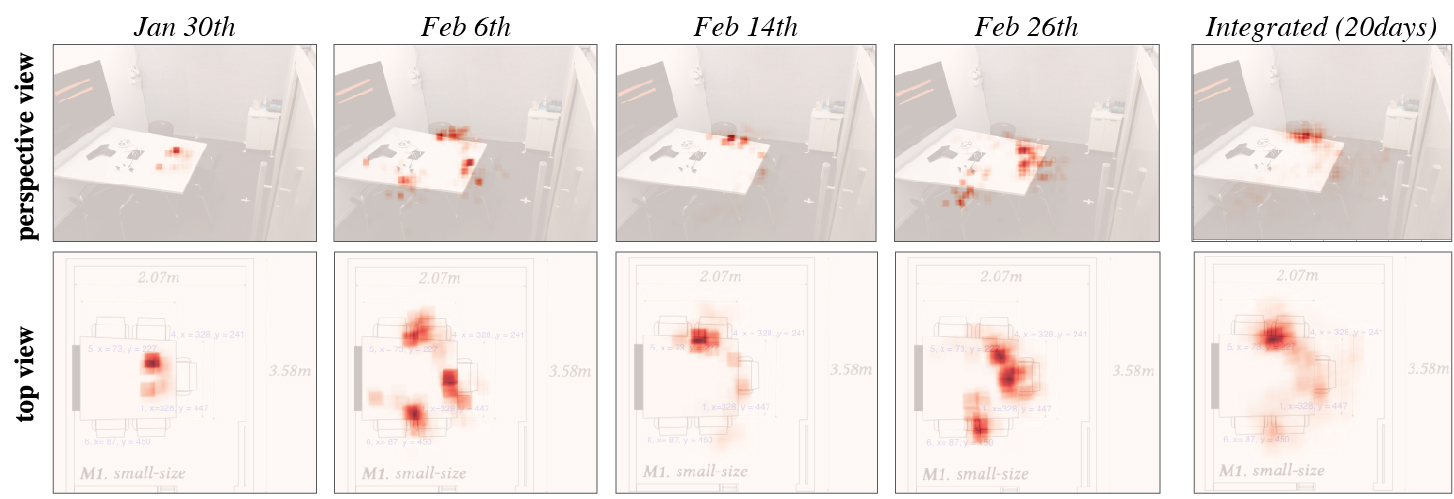}
        \caption{Desk contact heat-map using 2 monocular cameras (Top view and perspective view). The last column shows the accumulated contact results.}
        \label{fig:heatmap}
    \end{figure}

    \subsection{Desk usage indicator:}
    Heat-maps of selected days are generated by collecting the number of contacts being made for a point in the image. As shown in Figure \ref{fig:heatmap}, this observation is to show an overview of the desk usage. As we do not assign ids for the contacts, this means that for an area when $n$ numbers of the contacts being made by one person, it will still be counted as $n$ contacts. In the figures, we transformed the contact points to the desk model from the top-down view.

    \subsection{Office deck occupancy detection: }
\begin{figure}[t]
    \centering
    \includegraphics[width=0.99\columnwidth]{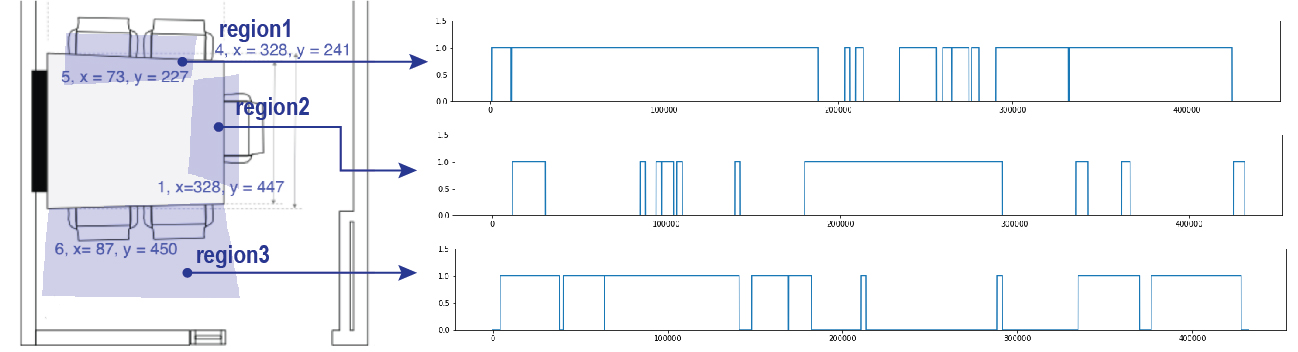}
        \caption{Region occupancy detection, figure on the left shows the region being monitored, and plots on the right is the occupancy indicator shows as time series data, each temporal unit is 0.04 second }
        \label{fig:occupancy}
    \end{figure}

    Most tracking algorithms \cite{zhan2020simple} \cite{zhou2019objects} \cite{zhou2020tracking} do not deal with long time disappearance of the object, which happens a lot when people are in the office with their body parts occluded by desks or chairs. Also, track-by-detection based algorithms will fail as the center of the tracking boxes vary by the occlusions. As a result, when a single person is sitting by the desk for an hour, hundreds of ids will be generated from that person, and this makes tracking algorithms very unstable for desk space occupancy detection task. We monitor the moment the desk regions are being contacted,  as shown in figure \ref{fig:occupancy}.  We assume that the desk space will not be reoccupied within a minute and for the same region, then if there is a gap of the usage less than a minute, we fill in the gap.



\section{Discussion and Future Work}

In this paper, we have discovered ways to find the contacted area on the surface of a static object to help people better understanding the COVID-19 spread path. We try to show people the insights of how complicated tasks for this application might be converted into very simple computer vision tasks.  
One of the interesting extensions is to apply this technique to movable objects. The projection constancy still holds if more dynamic mappings can be established.
Since the method we proposed can be used to label the data, another possible extension is to use the contact data as a training data set for a deep learning model that detects, from only a single camera view, when and where contact is being made by a human.

 \bibliographystyle{splncs04}
\bibliography{egbib}

\begin{thebibliography}{10}
\providecommand{\url}[1]{\texttt{#1}}
\providecommand{\urlprefix}{URL }
\providecommand{\doi}[1]{https://doi.org/#1}

\bibitem{agarwal2007high}
Agarwal, A., Izadi, S., Chandraker, M., Blake, A.: High precision multi-touch
  sensing on surfaces using overhead cameras. In: Second Annual IEEE
  International Workshop on Horizontal Interactive Human-Computer Systems
  (TABLETOP'07). pp. 197--200. IEEE (2007)

\bibitem{bulat2016human}
Bulat, A., Tzimiropoulos, G.: Human pose estimation via convolutional part
  heatmap regression. In: European Conference on Computer Vision. pp. 717--732.
  Springer (2016)

\bibitem{cao2018openpose}
Cao, Z., Hidalgo, G., Simon, T., Wei, S.E., Sheikh, Y.: Openpose: realtime
  multi-person 2d pose estimation using part affinity fields. arXiv preprint
  arXiv:1812.08008  (2018)

\bibitem{carreira2017quo}
Carreira, J., Zisserman, A.: Quo vadis, action recognition? a new model and the
  kinetics dataset. In: proceedings of the IEEE Conference on Computer Vision
  and Pattern Recognition. pp. 6299--6308 (2017)

\bibitem{chen2020action}
Chen, M.H., Li, B., Bao, Y., AlRegib, G., Kira, Z.: Action segmentation with
  joint self-supervised temporal domain adaptation. In: Proceedings of the
  IEEE/CVF Conference on Computer Vision and Pattern Recognition. pp.
  9454--9463 (2020)

\bibitem{fang2017rmpe}
Fang, H.S., Xie, S., Tai, Y.W., Lu, C.: Rmpe: Regional multi-person pose
  estimation. In: ICCV (2017)

\bibitem{farha2019ms}
Farha, Y.A., Gall, J.: Ms-tcn: Multi-stage temporal convolutional network for
  action segmentation. In: Proceedings of the IEEE Conference on Computer
  Vision and Pattern Recognition. pp. 3575--3584 (2019)

\bibitem{he2017mask}
He, K., Gkioxari, G., Doll{\'a}r, P., Girshick, R.: Mask r-cnn. In: Proceedings
  of the IEEE international conference on computer vision. pp. 2961--2969
  (2017)

\bibitem{ionescu2013human3}
Ionescu, C., Papava, D., Olaru, V., Sminchisescu, C.: Human3. 6m: Large scale
  datasets and predictive methods for 3d human sensing in natural environments.
  IEEE transactions on pattern analysis and machine intelligence
  \textbf{36}(7),  1325--1339 (2013)

\bibitem{kolotouros2019learning}
Kolotouros, N., Pavlakos, G., Black, M.J., Daniilidis, K.: Learning to
  reconstruct 3d human pose and shape via model-fitting in the loop. In:
  Proceedings of the IEEE International Conference on Computer Vision. pp.
  2252--2261 (2019)

\bibitem{lea2017temporal}
Lea, C., Flynn, M.D., Vidal, R., Reiter, A., Hager, G.D.: Temporal
  convolutional networks for action segmentation and detection. In: proceedings
  of the IEEE Conference on Computer Vision and Pattern Recognition. pp.
  156--165 (2017)

\bibitem{Lee:cscw:2019}
Lee, B., Lee, M., Zhang, P., Tessier, A., Khan, A.: An empirical study of how
  socio-spatial formations are influenced by interior elements and displays in
  an office context. Proc. ACM Hum.-Comput. Interact.  \textbf{3}(CSCW) (Nov
  2019). \doi{10.1145/3359160}, \url{https://doi.org/10.1145/3359160}

\bibitem{linqin2017dynamic}
Linqin, C., Shuangjie, C., Min, X., Jimin, Y., Jianrong, Z.: Dynamic hand
  gesture recognition using rgb-d data for natural human-computer interaction.
  Journal of Intelligent \& Fuzzy Systems  \textbf{32}(5),  3495--3507 (2017)

\bibitem{lowe2004distinctive}
Lowe, D.G.: Distinctive image features from scale-invariant keypoints.
  International journal of computer vision  \textbf{60}(2),  91--110 (2004)

\bibitem{lucas1981iterative}
Lucas, B.D., Kanade, T., et~al.: An iterative image registration technique with
  an application to stereo vision  (1981)

\bibitem{matsubara2017touch}
Matsubara, T., Mori, N., Niikura, T., Tano, S.: Touch detection method for
  non-display surface using multiple shadows of finger. In: 2017 IEEE 6th
  Global Conference on Consumer Electronics (GCCE). pp.~1--5. IEEE (2017)

\bibitem{pavllo20193d}
Pavllo, D., Feichtenhofer, C., Grangier, D., Auli, M.: 3d human pose estimation
  in video with temporal convolutions and semi-supervised training. In:
  Proceedings of the IEEE Conference on Computer Vision and Pattern
  Recognition. pp. 7753--7762 (2019)

\bibitem{suhr2009kanade}
Suhr, J.K.: Kanade-lucas-tomasi (klt) feature tracker. Computer Vision
  (EEE6503) pp. 9--18 (2009)

\bibitem{wan2015explore}
Wan, J., Guo, G., Li, S.Z.: Explore efficient local features from rgb-d data
  for one-shot learning gesture recognition. IEEE transactions on pattern
  analysis and machine intelligence  \textbf{38}(8),  1626--1639 (2015)

\bibitem{wei2016convolutional}
Wei, S.E., Ramakrishna, V., Kanade, T., Sheikh, Y.: Convolutional pose
  machines. In: Proceedings of the IEEE conference on Computer Vision and
  Pattern Recognition. pp. 4724--4732 (2016)

\bibitem{yan2018spatial}
Yan, S., Xiong, Y., Lin, D.: Spatial temporal graph convolutional networks for
  skeleton-based action recognition. In: Thirty-second AAAI conference on
  artificial intelligence (2018)

\bibitem{zhan2020simple}
Zhan, Y., Wang, C., Wang, X., Zeng, W., Liu, W.: A simple baseline for
  multi-object tracking. arXiv preprint arXiv:2004.01888  (2020)

\bibitem{zhou2020tracking}
Zhou, X., Koltun, V., Kr{\"a}henb{\"u}hl, P.: Tracking objects as points.
  arXiv:2004.01177  (2020)

\bibitem{zhou2019objects}
Zhou, X., Wang, D., Kr{\"a}henb{\"u}hl, P.: Objects as points. In: arXiv
  preprint arXiv:1904.07850 (2019)

\end{thebibliography}

\end{document}